\documentclass[letterpaper]{article} % DO NOT CHANGE THIS
\usepackage{aaai25}  % DO NOT CHANGE THIS
\usepackage{times}  % DO NOT CHANGE THIS
\usepackage{helvet}  % DO NOT CHANGE THIS
\usepackage{courier}  % DO NOT CHANGE THIS
\usepackage[hyphens]{url}  % DO NOT CHANGE THIS
\usepackage{graphicx} % DO NOT CHANGE THIS
\usepackage{natbib}  % DO NOT CHANGE THIS AND DO NOT ADD ANY OPTIONS TO IT
\usepackage{caption} % DO NOT CHANGE THIS AND DO NOT ADD ANY OPTIONS TO IT
\usepackage{algorithm}
\usepackage{algorithmic}

\usepackage{color,xcolor}
\usepackage{colortbl}
\newcommand{\pub}[1]{\color{gray}{\scriptsize{[{#1}]}}}

\usepackage{subcaption}
\usepackage{bm}
\usepackage{amssymb}
\usepackage{amsmath}
\DeclareMathOperator*{\argmax}{arg\,max}

\newcommand{\framework}{Audio-Language-Referenced SAM 2}
\newcommand{\framesimple}{AL-Ref-SAM 2}

\usepackage{newfloat}
\usepackage{listings}
\DeclareCaptionStyle{ruled}{labelfont=normalfont,labelsep=colon,strut=off} % DO NOT CHANGE THIS
\floatstyle{ruled}
\newfloat{listing}{tb}{lst}{}
\floatname{listing}{Listing}
\title{Unleashing the Temporal-Spatial Reasoning Capacity of GPT for Training-Free Audio and Language Referenced Video Object Segmentation}
\author {
    Shaofei Huang\textsuperscript{\rm 1,\rm 2}\thanks{These authors contributed equally. Shaofei Huang is a visiting scholar at Hefei University of Technology.} \quad
    Rui Ling\textsuperscript{\rm 3}\footnotemark[1] \quad
    Hongyu Li\textsuperscript{\rm 4}\footnotemark[1] \quad
    Tianrui Hui\textsuperscript{\rm 1}\thanks{Tianrui Hui and Si Liu are the corresponding authors.} \\
    Zongheng Tang\textsuperscript{\rm 4} \quad
    Xiaoming Wei\textsuperscript{\rm 5} \quad
    Jizhong Han\textsuperscript{\rm 2} \quad
    Si Liu\textsuperscript{\rm 4}\footnotemark[2]
}
\affiliations {
    \textsuperscript{\rm 1}School of Computer Science and Information Engineering, Hefei University of Technology \\
    \textsuperscript{\rm 2}Institute of Information Engineering, Chinese Academy of Sciences\\
    \textsuperscript{\rm 3}School of Computer Science and Engineering, Beihang University\\
    \textsuperscript{\rm 4}School of Artificial Intelligence, Beihang University \quad
    \textsuperscript{\rm 5}Meituan\\
    {\tt\small \{nowherespyfly,~hongyuer0317,~huitianrui\}@gmail.com \quad weixiaoming@meituan.com}
    {\tt\small \{20373184,~tzhhhh123,~liusi\}@buaa.edu.cn \quad hanjizhong@iie.ac.cn}
}

\begin{document}

\maketitle

\begin{abstract}
In this paper, we propose an \framework~(\framesimple) pipeline to explore the training-free paradigm for audio and language-referenced video object segmentation, namely AVS and RVOS tasks.
The intuitive solution leverages GroundingDINO to identify the target object from a single frame and SAM 2 to segment the identified object throughout the video, which is less robust to spatiotemporal variations due to a lack of video context exploration.
Thus, in our \framesimple~pipeline, we propose a novel GPT-assisted Pivot Selection (GPT-PS) module to instruct GPT-4 to perform two-step temporal-spatial reasoning for sequentially selecting pivot frames and pivot boxes, thereby providing SAM 2 with a high-quality initial object prompt.
Within GPT-PS, two task-specific Chain-of-Thought prompts are designed to unleash GPT's temporal-spatial reasoning capacity by guiding GPT to make selections based on a comprehensive understanding of video and reference information.
Furthermore, we propose a Language-Binded Reference Unification (LBRU) module to convert audio signals into language-formatted references, thereby unifying the formats of AVS and RVOS tasks in the same pipeline.
Extensive experiments show that our training-free \framesimple~pipeline achieves performances comparable to or even better than fully-supervised fine-tuning methods.
\end{abstract}

% Uncomment the following to link to your code, datasets, an extended version or similar.
\begin{links}
    \link{Code}{https://github.com/appletea233/AL-Ref-SAM2}
\end{links}

\section{Introduction}

\begin{figure}[!t]
\centering
\includegraphics[width=0.95\linewidth]{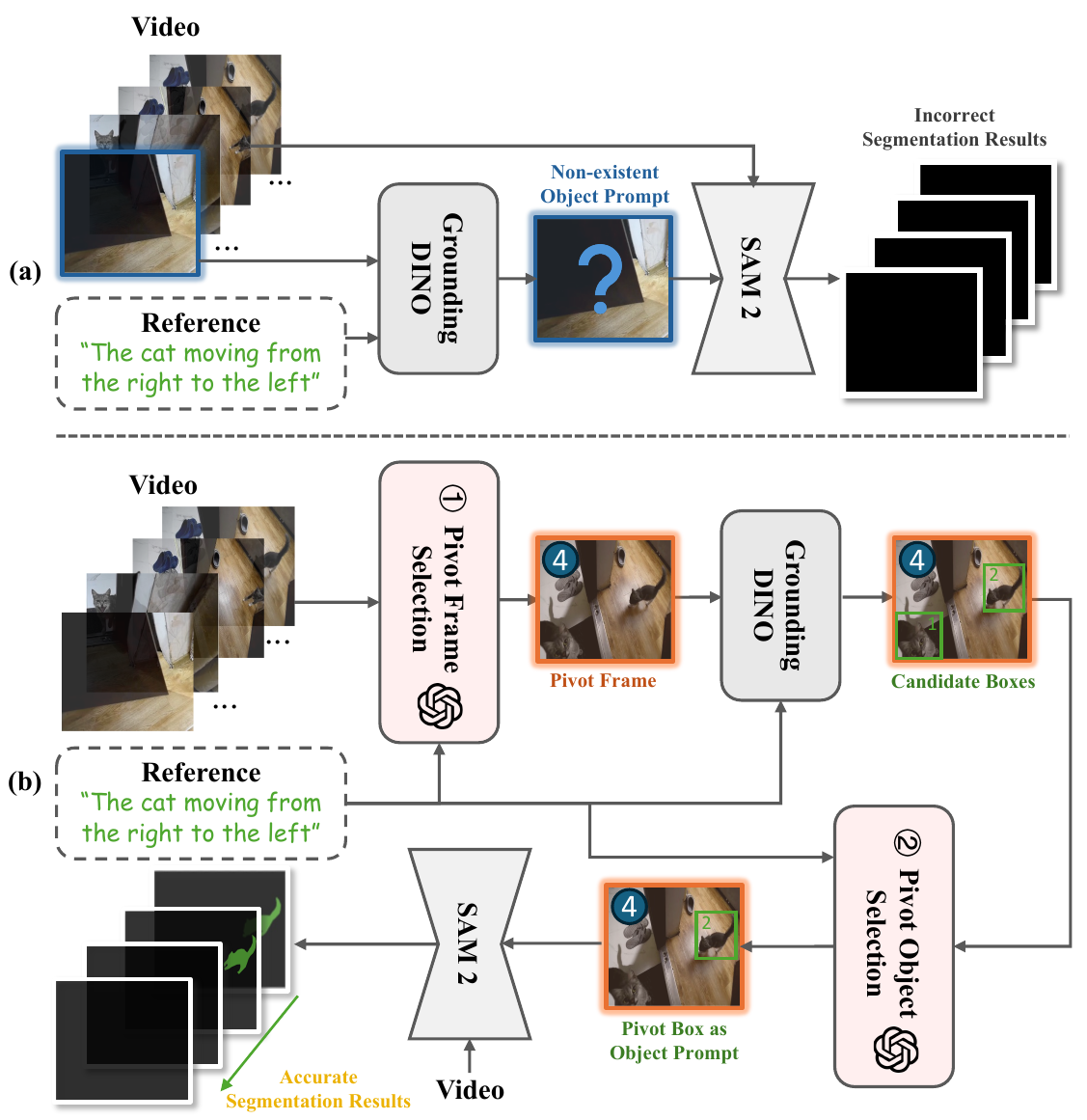}
\caption{(a) For the training-free baseline, naively choosing the first frame to generate the object prompt may yield completely wrong predictions on the video since the first frame does not contain any relevant object. (b) Our method leverages GPT-4 to perform two-step temporal-spatial reasoning, selecting the frame and box that best reflects the reference information. The selected box serves as a more accurate object prompt to SAM 2 for better segmentation results.}
\label{fig:intro}
\end{figure}

Video Object Segmentation (VOS), which involves segmenting and tracking specific objects throughout a video sequence, has garnered growing attention due to its significant potential in various real-world applications.
Recent works further incorporate additional multimodal reference information, such as language and audio, making it more convenient and flexible to locate objects of interest in videos.
For example, Referring Video Object Segmentation (RVOS)~\cite{seo2020urvos} specifies the target object through language description, while Audio Visual Segmentation (AVS)~\cite{zhou2022audio} segments objects emitting the sounds in the associated audio.
However, most existing methods for these two tasks require extensive training in a fully-supervised manner, which suffers from high training costs and labor-intensive label annotations.
Therefore, in this work, we propose to explore the training-free paradigm for multimodal information-aided VOS tasks.

Despite their cost-effectiveness and simplicity properties, training-free methods still exhibit significantly inferior performance compared with finetuning-based ones, owing to their lack of adaptability to the data distribution of specific VOS tasks.
Recently, foundation models~\cite{radford2021learning,achiam2023gpt,liu2024visual,kirillov2023segment} have demonstrated strong generalization abilities across various tasks, illuminating a promising path to bridging this performance gap.
For instance, SAM 2~\cite{ravi2024sam} shows powerful zero-shot transferability for promptable video segmentation, while GroundingDINO~\cite{liu2023grounding} excels at locating the target objects in a single image based on language descriptions.
Leveraging these foundation models, an intuitive three-stage pipeline for training-free multimodal VOS tasks can be constructed~\cite{ren2024grounded}: (1) extracting reference information from the multimodal input, (2) identifying the target object in the initial frame using GroundingDINO according to the extracted reference, and (3) segmenting the identified target object throughout the entire video using SAM 2.
However, since GroundingDINO is designed for single-image grounding tasks, this intuitive solution may fail to identify target objects effectively due to its lack of video context exploration, making it less robust to spatiotemporal variations. 
As shown in Figure~\ref{fig:intro}(a), if the target object is inaccurately identified or non-existent in the initial frame, subsequent frames may yield degraded segmentation results due to the incorrect object prompt to SAM 2.

To alleviate this limitation, we propose a novel two-step temporal-spatial reasoning flow that first selects a \textbf{\textit{pivot frame}} from the entire video via temporal reasoning, and then identifies a \textbf{\textit{pivot box}} from multiple candidate boxes on the pivot frame via spatial reasoning, thus providing SAM 2 with an accurate object prompt for initiating video segmentation.
As shown in Figure~\ref{fig:intro}(b), the pivot frame is defined as the specific frame where the target object (\textit{i.e.}, the referent) clearly appears without being occluded or blurred, while the pivot box refers to the box on the pivot frame that best matches the reference information, \textit{i.e.}, the referent's box.
To implement this complex reasoning process within a training-free framework, we incorporate GPT-4~\cite{achiam2023gpt} to leverage its vision-language comprehension and reasoning capacity.
However, temporal and spatial reasoning are both inherently complex, necessitating an exhaustive comprehension of video and reference information.
Naively providing GPT with abstract instructions such as \textit{``select the frame where the target object appears clearly''} or \textit{``locate the object according to the reference''} often fails to yield satisfactory results, as it typically chooses a middle frame as the pivot frame or the most salient object as the pivot box.
We attribute this phenomenon to the logical shortcuts that GPT tends to take when confronted with complex reasoning tasks, often opting for the path of least resistance and resulting in superficial answers in the absence of explicit guidance~\cite{wei2022chain}.
To tackle this issue, we meticulously design two task-specific Chain-of-Thought (CoT) prompts for the aforementioned two steps respectively, guiding GPT step-by-step to first form a comprehensive understanding of the video-reference pairs before answering the final questions.
We implement the two steps as a GPT-assisted Pivot Selection (GPT-PS) module which unleashes the temporal-spatial reasoning capacity of GPT, prompting it to perform complex reasoning based on visual and reference information to make accurate judgments.

Additionally, given that audio signals are semantically ambiguous and intrinsically redundant, unifying audio into language format benefits more in the training-free setting for AVS task.
To this end, we design a Language-Binded Reference Unification (LBRU) module to convert the audio signal into descriptions of the sounding objects from an acoustic perspective.
Concretely, to eliminate the semantic ambiguity of audio signals and the interference of background noise, we incorporate visual context and leverage GPT-4 to identify the categories of the sounding objects from audio-video pairs.
Symbolic representation is employed to encode video and audio data into sequences recognizable by GPT.
By converting audio signals into higher-level, clearly defined language-formatted references, we can not only mitigate the inferior performance caused by the inherent shortcomings of audio, but also unify the formats of AVS and RVOS tasks, enabling our pipeline to handle both seamlessly.
Integrating LBRU and GPT-PS modules, we name our pipeline as \framework ~(\framesimple)~which performs training-free unified audio- and language-referenced video object segmentation.

The contributions of our paper are summarized as follows:
(1) We propose a GPT-assisted Pivot Selection (GPT-PS) module where GPT-4 is instructed to perform two-step temporal-spatial reasoning for selecting pivot frames and boxes that match with references, providing high-quality prompts to SAM 2 for precise video segmentation.
(2) We propose a Language-Binded Reference Unification (LBRU) module that converts audio signals into language-formatted references, unifying AVS and RVOS tasks to be handled in the same pipeline.
(3) Extensive experiments demonstrate that our training-free \framesimple~ pipeline achieves results that are comparable to, and in certain benchmarks even better than, those of finetuning-based methods on both tasks.

\begin{figure*}[t]
\centering
\includegraphics[width=0.88\linewidth]{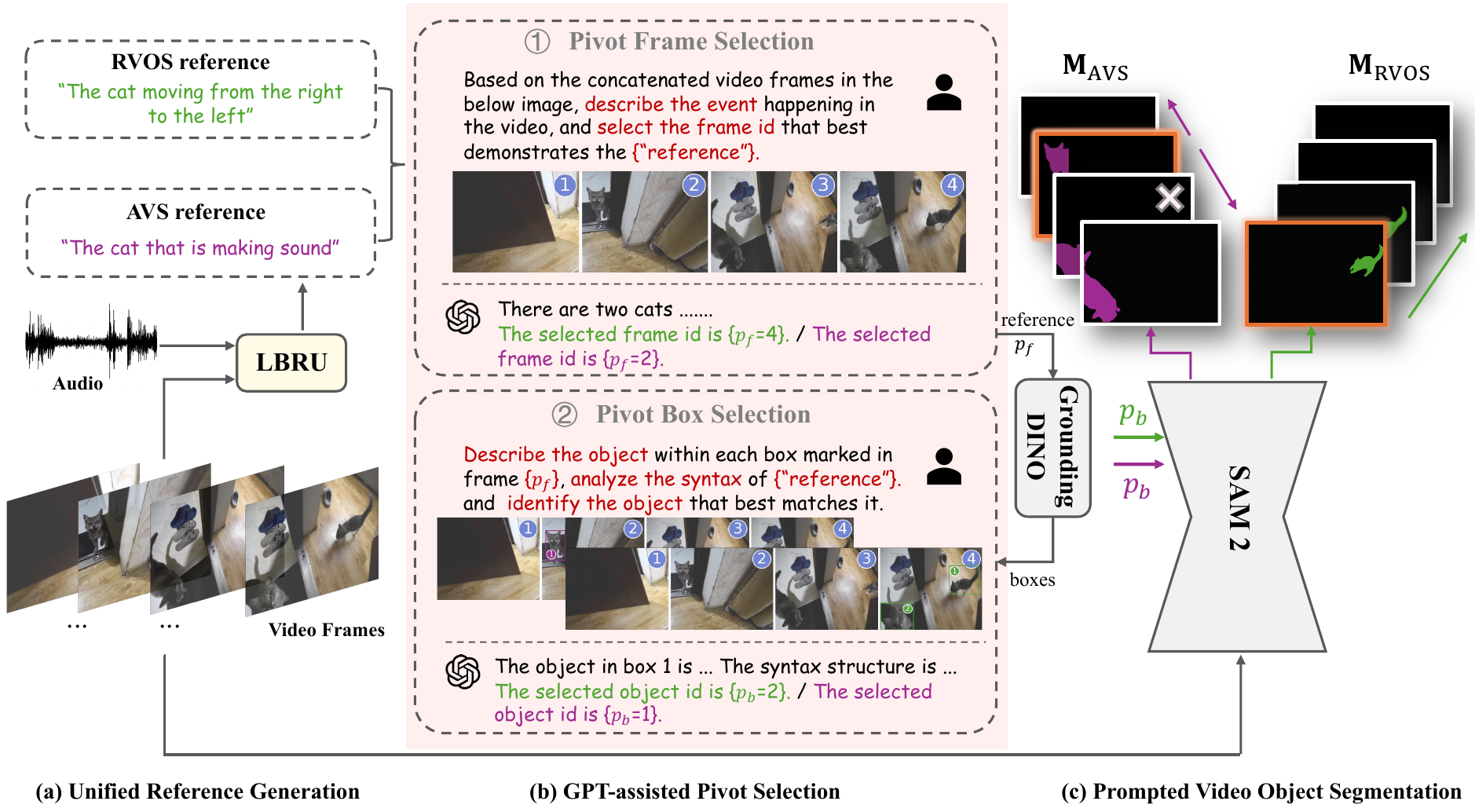}
\caption{The overall pipeline of our proposed \framework. (a) Generate a language-formatted reference that specifies the objects to be segmented for both RVOS and AVS tasks. (b) Select the pivot frame and pivot box through two-step temporal-spatial reasoning. (c) Prompt SAM 2 with the selected pivot box to obtain the mask sequence of the target object across the entire video. The symbol $\times$ on $\mathbf{M}_{\rm{AVS}}$ represents the mask to be filtered out where the sound-emitting object is silent. \textit{Different colors are used to denote the data flow of RVOS and AVS tasks respectively.}}
\label{fig:pipeline}
\end{figure*}

\section{Related Works}

\subsection{Referring Video Object Segmentation}
Different from Referring Image Segmentation~\cite{hui2020linguistic,liu2021cross,huang2020referring,liu2024primitivenet}, Referring Video Object Segmentation (RVOS)~\cite{seo2020urvos,hui2021collaborative,ding2022language,hui2023language} aims to segment the target object matched with the description of a given sentence.
ReferFormer~\cite{wu2022language} and MTTR~\cite{botach2022end} utilize language as queries which attend to the relevant visual regions through Transformer models.
Temporal information is also investigated to align object motion with language expression by reasoning across multiple temporal scales~\cite{han2023html,tang2021human} or temporal interaction between the global referent token and object queries~\cite{tang2023temporal}.
LoSh~\cite{yuan2024losh} proposes to utilize both long and short text expressions to understand the appearance and motion cues of the target object.
In this paper, we tackle the RVOS task in the training-free setting by leveraging the temporal-spatial reasoning capacity of GPT.

\subsection{Audio-Visual Segmentation}
The goal of Audio-Visual Segmentation~\cite{zhou2022audio} is to localize the objects emitting the given sounds in a video by pixel-level masks and predict their semantic category labels~\cite{zhou2023audio} as well.
AQFormer~\cite{huang2023discovering} constructs a set of object queries conditioned on audio information to associate visual regions of sounding objects with audio cues.
\cite{chen2024unraveling} propose audio-visual supervised contrastive learning with an informative sample mining technique, aiming to utilize discriminative contrastive samples to strengthen cross-modal understanding.
\cite{liu2024audio} propose leveraging motion cues from neighboring frames and semantic cues from distant frames to extract information from abundant unlabeled frames for improving performance.
In this paper, considering the inherent complexity and ambiguity of audio, we express it as a form of language and leverage foundation models to achieve unified and training-free video object segmentation.

\subsection{Foundation Models}
Foundation models are large, pre-trained models that serve as a general-purpose base for a wide range of downstream tasks across different domains~\cite{shi2024mask,liu2021human,wang2023transferring}.
GPT-4~\cite{achiam2023gpt} has garnered significant attention for its impressive conversational and reasoning capabilities.
In addition, other open-source vision-language models (such as CLIP~\cite{radford2021learning}, LLaVA~\cite{liu2024visual}, OPT~\cite{zhang2022opt}, Diffusion Models~\cite{rombach2022high,zhang2023controlvideo}, \textit{etc}.) have also demonstrated excellent multimodal understanding and generation abilities.
In terms of visual foundation models, the well-known SAM~\cite{kirillov2023segment}, trained on over a billion masks, demonstrates powerful zero-shot image segmentation capabilities based on various prompts.
SAM-Track~\cite{cheng2023segment} and SAM 2~\cite{ravi2024sam} further extend the fine-grained perception ability of SAM to videos, enabling general segmentation and tracking of objects within video sequences.
In this paper, we design a novel two-step temporal-spatial reasoning flow to exploit the strong task execution capability of foundation models for training-free multimodal video object segmentation.

\section{\framework}

\subsection{Pipeline Overview}
The overall architecture of our \framework~pipeline (\framesimple) in illustrated in Figure~\ref{fig:pipeline}.
Given a video clip consisting of $T$ frames, and either a language description or an audio clip specifying the objects to be segmented (termed the \textit{referent}), the goal of our \framesimple~ pipeline is to obtain the mask sequence $\mathbf{M}_{\rm{RVOS}}$ or $\mathbf{M}_{\rm{AVS}} \in \mathbb{R}^{T\times H\times W}$ of the referent across the whole video. 
Here, $H$ and $W$ denote the height and width of the video frames respectively.
The language description is utilized in the RVOS task as the reference for indicating the referent.
For the AVS task, we feed both the audio and video clips to the Language-Binded Reference Unification (LBRU) module to acquire a reference that describes the sounding objects from the acoustic aspect, thereby unifying the format of the AVS reference with that of the RVOS reference.
Subsequently, the obtained reference and the video clip are processed by the GPT-assisted Pivot Selection (GPT-PS) module to identify a high-quality bounding box of the referent in a specific frame where the referent clearly appears.
Finally, the selected bounding box serves as the pivot box to effectively prompt SAM 2 to segment the referent and propagate its mask forward and backward through the entire video clip.
For the AVS task, we use sound event detection to segment the audio clip and filter out silent frames.

\subsection{Language-Binded Reference Unification}
\label{sec:lbru}

AVS aims to segment the sounding objects within the video clip based on its corresponding audio content, which can also be formulated as using the reference of ``\textit{the} [OBJ] \textit{that is making sound}'' for segmentation, where ``[OBJ]'' denotes the category of the specific sounding object in the video.
In this way, the audio clip in the AVS task can be converted into a language-formatted reference for facilitating better task unification.
To obtain the categories of all the sounding objects, an intuitive approach involves applying an audio classifier (\textit{e.g.}, BEATs~\cite{chen2022beats}) to the audio clip, classifying it into several categories.
However, due to the presence of background noise and the ambiguity of audio information, this audio-only approach may collect incorrect or unnecessary object categories, leading to suboptimal segmentation performance.
Therefore, we incorporate visual context and leverage the inherent vision-language understanding capabilities of MLLMs (\textit{e.g.}, GPT-4) to accurately identify the categories of the actual sounding objects present in the video.

\begin{figure}[!htbp]
\centering
\includegraphics[width=0.9\linewidth]{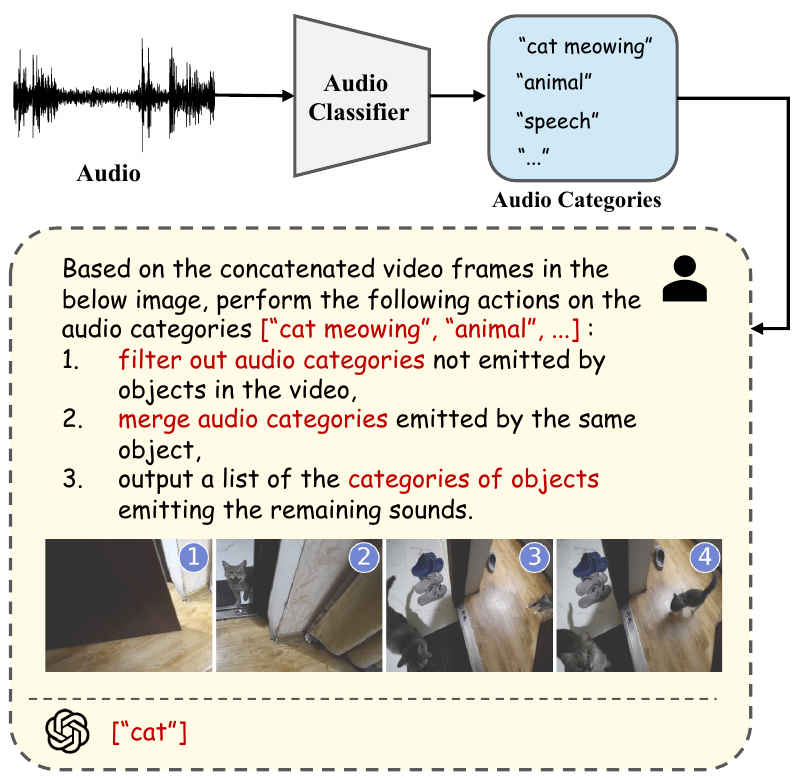}
\caption{Detailed illustration of our Language-Binded Reference Unification module.}
\label{fig:lbru}
\end{figure}

Given that GPT-4 is currently unable to comprehend audio and video data, we first encode the audio and video data into GPT-recognizable sequences through symbolic representation.
Specifically, as shown in Figure~\ref{fig:lbru}, for the audio clip, we feed it into the audio classifier and retain the categories with the top-$k$ classification confidence scores.
The text of the retained categories is then organized into a list as ${\rm \mathbf{X}_a} = [{\rm CLS}_1, {\rm CLS}_2, ..., {\rm CLS}_k]$.
As for the video clip, we evenly sample $m$ frames and concatenate them sequentially into a single image, with the frame ID marked on each frame.
The concatenated image is denoted as $\rm \mathbf{X}_v$.
We then combine the above symbolic representations with our carefully curated language command $\rm \mathbf{X}_l$ as the prompt $\rm \mathbf{X}_{pa}$ to GPT: $\rm \mathbf{X}_{pa}=[\mathbf{X}_l, \mathbf{X}_a, \mathbf{X}_v]$.
The goal of $\rm \mathbf{X}_{pa}$ is to guide GPT step by step in filtering and merging audio labels based on the content of the video frames, ultimately outputting the categories of objects emitting the remaining sounds.
These categories are denoted as $[\mathrm{OBJ}_1, \mathrm{OBJ}_2..., \mathrm{OBJ}_n]$, where $n$ represents the total number of sounding object categories in the video.
We convert each category to a separate reference and process it individually.

\subsection{GPT-Assisted Pivot-Selection}
\label{sec:gpt-ps}
Our GPT-assisted Pivot Selection (GPT-PS) module aims to leverage GPT's vision-language reasoning capabilities to obtain a highly distinguishable bounding box of the referent, thus prompting SAM 2 for precise segmentation.
The reasoning process involves two steps.
In the first step, temporal reasoning is conducted to select a pivot frame from the entire video, in which the referent can be clearly distinguished.
Subsequently, spatial reasoning is carried out on the pivot frame to select the pivot box that best matches the reference description from multiple candidates within this frame.
Utilizing the pivot frame as the starting point for SAM 2's segmentation process, we propagate the object mask within the pivot box forward and backward throughout the entire video to obtain the complete mask sequence of the referent.

\textbf{Step 1: Pivot Frame Selection.}
As illustrated in the upper part of Figure~\ref{fig:pipeline}(b), we employ the same way as described in LBRU~\ref{sec:lbru} to obtain the symbolic representation $\rm \mathbf{X}_v$ of video information.
The colored frame numbers marked on $\rm \mathbf{X}_v$ can inform GPT of the order of the concatenated frames, thereby enhancing its comprehension of the video information.
Pivot frame selection is a complex temporal reasoning process that necessitates GPT to both comprehend the events occurring in the video and preliminarily identify the actual object mentioned in the reference.
To this end, we design the Pivot-Frame CoT prompt, $\rm \mathbf{X}_{pf}$, which instructs GPT to first describe the temporal event for enhancing its comprehensive understanding of the video content, and then select the pivot frame from the sampled frames in $\rm \mathbf{X}_v$ where the reference content is most easily recognized.
The obtained frame ID is denoted as $p_f$.

\begin{table*}[t!]
\setlength{\tabcolsep}{7.8pt}
\centering
\footnotesize
\begin{tabular}{l| c |c c c | c c c | c c c}
\rowcolor[gray]{.9} \hline
&  &  \multicolumn{3}{c |}{Ref-YouTube-VOS} & \multicolumn{3}{c|}{Ref-DAVIS17}  & \multicolumn{3}{c}{MeViS} \\
\rowcolor[gray]{.9} 
Method &Reference & \( \mathcal{J} \)\&\( \mathcal{F} \) & \( \mathcal{J} \) & \( \mathcal{F} \)  &  \( \mathcal{J} \)\&\( \mathcal{F} \) & \( \mathcal{J} \) & \( \mathcal{F} \) &  \( \mathcal{J} \)\&\( \mathcal{F} \) & \( \mathcal{J} \) & \( \mathcal{F} \) \\ \hline\hline
\multicolumn{11}{c}{Fully-Supervised Fine-Tuning} \\ \hline
ReferFormer~\shortcite{wu2022language} & \pub{CVPR'22} & 62.9 & 61.3 & 64.6 & 61.1 & 58.1 & 64.1 & 31.0 & 29.8 & 32.2 \\
OnlineRefer~\shortcite{wu2023onlinerefer} & \pub{ICCV'23} &62.9& 61.0 &64.7 &62.4 &59.1 &65.6 & - & - & -\\
HTML~\shortcite{han2023html} &\pub{ICCV'23} & 63.4 &61.5 &65.2 &62.1 &59.2 &65.1 & - & - & -\\
SgMg~\shortcite{miao2023spectrum}  & \pub{ICCV'23} &  {65.7} & {63.9} & {67.4}  & {63.3} & {60.6} & {66.0} & - & - & - \\ 
LMPM~\shortcite{ding2023mevis} & \pub{ICCV'23} & - & - & - & - & - & - & 37.2 & 34.2 & 40.2 \\
TempCD~\shortcite{tang2023temporal}&\pub{ICCV'23} &65.8&63.6 &68.0 & 64.6& 61.6& 67.6 & - & - & -\\
SOC~\shortcite{luo2024soc} & \pub{NIPS'23} &66.0 &64.1& 67.9& 64.2 &61.0 &67.4 & - & - & -\\
VD-IT~\shortcite{zhu2024exploring}&\pub{ECCV'24} &66.5&64.4 &68.5 & 63.0& 59.9& 66.1 & - & - & -\\
LoSh ~\shortcite{yuan2024losh} &\pub{CVPR'24}& 67.2 & 65.4& 69.0 & 64.3 & 61.8 & 66.8& - & - & - \\
DsHmp ~\shortcite{he2024decoupling} &\pub{CVPR'24}& 67.1 & 65.0& 69.1 & 64.9 & 61.7 & 68.1 & 46.4 & 43.0 & 49.8 \\ \hline
\multicolumn{11}{c}{Weakly-Supervised Fine-Tuning} \\ \hline
WRVOS~\shortcite{zhao2023learning} & \pub{arXiv'23} & 46.6 & 45.6 & 47.6 & 47.3 & 44.6 & 50.0 & - & - & - \\
GroPrompt~\shortcite{lin2024groprompt} & \pub{CVPRW'24} & 65.5 & 64.1 & 66.9 & 70.6 & 67.8 & 73.3 & - & - & - \\ \hline
\multicolumn{11}{c}{\cellcolor{red!5} Training-Free} \\ \hline
\rowcolor{red!5}G-L + SAM 2 ~\shortcite{yu2023zero}&\pub{CVPR'23}& 27.0 & 24.3 & 29.7 & 40.6 & 37.6& 43.6 & 23.7 & 20.4 & 30.0 \\
\rowcolor{red!5}G-L (SAM) + SAM 2 ~\shortcite{yu2023zero}&\pub{CVPR'23}& 33.6 & 29.9 & 37.3 & 46.9 & 44.0 & 49.7 & 26.6 & 22.7 & 30.5  \\
\rowcolor{red!5} Grounded-SAM~\shortcite{ren2024grounded}* &\pub{arXiv'24}& 62.3 & 61.0 & 63.6 & 65.2 & 62.3 & 68.0 & - & - & -\\
\rowcolor{red!5} Grounded-SAM 2~\shortcite{ren2024grounded}$\dagger$ &\pub{arXiv'24} & 64.8 &62.5& 67.0& 66.2 &62.6 &69.7 & 38.9 & 35.7 & 42.1\\
\rowcolor{red!15} \textbf{\framesimple~(Ours)} & - &\textbf{67.9}& \textbf{65.9}& \textbf{69.9}&\textbf{74.2} &\textbf{70.4} &\textbf{78.0} &\textbf{42.8} &\textbf{39.5} &\textbf{46.2} \\ \hline
\end{tabular}
\caption{Comparison with state-of-the-art methods on the validation sets of Ref-YouTube-VOS, Ref-DAVIS17 and MeViS datasets. * Results are adopted from GroPrompt. $\dagger$ Results are obtained by running the official code.}
\label{tab:ytvos_davis}
\end{table*}

\begin{table*}[t!]
\setlength{\tabcolsep}{10pt}
\centering
\footnotesize
\begin{tabular}{l | c | c c | c c | c c | c c}
\rowcolor[gray]{.9} \hline
&  &  \multicolumn{2}{c |}{S4} & \multicolumn{2}{c |}{MS3} & \multicolumn{2}{c |}{AVSS} & \multicolumn{2}{c}{AVSS-V2-Binary} \\
\rowcolor[gray]{.9}
Method &Reference & \( \mathcal{M_J} \) & \( \mathcal{M_F} \) & \( \mathcal{M_J} \) & \( \mathcal{M_F} \) & \( \mathcal{M_J} \) & \( \mathcal{M_F} \) & \( \mathcal{M_J} \) & \( \mathcal{M_F} \) \\ \hline\hline
\multicolumn{10}{c}{Fully-Supervised Fine-Tuning} \\ \hline
TPAVI~\shortcite{zhou2022audio,zhou2023audio} & \pub{ECCV'22} & 78.7 & 87.9 & 54.0 & 64.5 &29.8 & 35.2 & 62.5 & 75.6 \\
AQFormer~\shortcite{huang2023discovering} & \pub{IJCAI'23} & 81.6 & 89.4 & 61.1 & 72.1 & - & - & - & - \\
CATR~\shortcite{li2023catr} & \pub{ACM MM'23} & 81.4 & 89.6 & 59.0 & 70.0 & 32.8 & 38.5 & - & - \\
BAVS~\shortcite{liu2024bavs} & \pub{TMM'24} & 82.0 & 88.6 & 58.6 & 65.5 & 32.6 & 36.4 & - & - \\
Audio-SAM~\shortcite{wang2024prompting} & \pub{AAAI'24} & 56.3 & 72.7 & 33.7 & 45.9 & - & - & 57.4 & 68.4 \\
SAM-Fusion~\shortcite{wang2024prompting} & \pub{AAAI'24} & 71.9 & 77.5 & 50.6 & 63.7 & - & - & 60.2 & 72.4 \\
GAVS~\shortcite{wang2024prompting} & \pub{AAAI'24} & 80.1 & 90.2 & 63.7 & 77.4 & - & - & 67.7 & 78.8 \\
AVSegFormer~\shortcite{gao2024avsegformer} & \pub{AAAI'24} & 82.1 & 89.9 & 58.4 & 69.3 & 36.7 & 42.0 & 64.3 & 75.9 \\
COMBO~\shortcite{yang2024cooperation} & \pub{CVPR'24} & 84.7 & 91.9 & 59.2 & 71.2 & 42.1 & 46.1 & - & - \\ \hline
\multicolumn{10}{c}{Weakly-Supervised Fine-Tuning} \\ \hline
WS-AVS~\shortcite{mo2024weakly} & \pub{NeurIPS'24} & 34.1 & 51.8 & 30.9 & 46.9 & - & - & - & - \\
MoCA~\shortcite{bhosale2024unsupervised} & \pub{arXiv'24} & 68.0 & 79.0 & 57.0 & 62.0 & 31.0 & 33.0 & - & - \\ \hline
\multicolumn{10}{c}{\cellcolor{red!5}Training-Free} \\ \hline
\rowcolor{red!5}AT-GDINO-SAM~\shortcite{bhosale2024unsupervised} & \pub{arXiv'24} & 38.0 & 46.0 & 25.0 & 29.0 & 24.0 & 25.0 & - & - \\
\rowcolor{red!5}SAM-BIND~\shortcite{bhosale2024unsupervised} & \pub{arXiv'24} & 42.0 & 51.0 & 28.0 & 36.0 & 24.0 & 26.0 & - & - \\
\rowcolor{red!5}OWOD-BIND~\shortcite{bhosale2024unsupervised} & \pub{arXiv'24} & 58.0 & 67.0 & 34.0 & 44.0 & 26.0 & 29.0 & - & - \\
\rowcolor{red!15} \textbf{\framesimple~(Ours)} & - &\textbf{70.5} &\textbf{81.1} &\textbf{48.6} &\textbf{53.5} &\textbf{36.0} &\textbf{39.8} &\textbf{59.2} &\textbf{66.2} \\ \hline
\end{tabular}
\caption{Comparison with state-of-the-art methods on the different subsets of the AVSBench dataset. ``.0'' in the results of other training-free methods is due to the rounding errors from their original papers.}
\label{tab:avs}
\end{table*}

\textbf{Step 2: Pivot Box Selection.}
As illustrated in the lower part of Figure~\ref{fig:pipeline}(b), we first employ GroundingDINO to predict candidate boxes of the referent on the pivot frame according to the reference description.
Since GroundingDINO is designed for grounding tasks on a single image, whereas our reference contains extensive temporal-related information, it is challenging to accurately predict the target object based solely on a single frame.
Thus, we lower the confidence threshold for GroundingDINO to generate multiple candidates, maximizing the likelihood of the inclusion of the actual referent.
Subsequently, we paint the candidate boxes on the pivot frame and mark each with a box ID to designate the regions for GPT's attention.
We also sequentially concatenate the other sampled frames with the marked pivot frame to provide temporal context for distinguishing the referent from multiple candidates.
To further aid this process, a Pivot-Box CoT prompt, $\rm \mathbf{X}_{pb}$, is designed to guide GPT step-by-step in reasoning about the pivot box that best matches the reference description on the pivot frame.
Specifically, based on the prior knowledge of the video event obtained in step 1, $\rm \mathbf{X}_{pb}$ first instructs GPT to describe the objects in each box, enabling it to comprehend candidates' appearance, motion, and interrelationships.
Afterward, GPT is required to perform syntactic analysis of the reference description to accurately identify the subject being referred to.
Finally, the candidate box that best corresponds to the description of the identified subject is selected as the pivot box $p_b$.
For illustrative purposes, we present an abridged version of the prompts in Figure~\ref{fig:pipeline}(b).

\section{Experiments}
\subsection{Datasets and Evaluation Metrics}
We adopt Ref-YouTube-VOS~\cite{seo2020urvos}, Ref-DAVIS17~\cite{khoreva2019video}, MeViS~\cite{ding2023mevis} for RVOS evaluation, and AVSBench~\cite{zhou2022audio} datasets for AVS evaluation respectively.
In terms of evaluation metrics, we adopt region similarity $\mathcal{J}$ (average IoU), contour accuracy $\mathcal{F}$ and their average $\mathcal{J}\&\mathcal{F}$ for RVOS.
The metrics $\mathcal{M_J}$ and $\mathcal{M_F}$ for AVS are the same as $\mathcal{J}$ and $\mathcal{F}$.

\subsection{Implementation Details}
We adopt the $\texttt{sam2\_hiera\_large}$ version of SAM 2 as the video segmentor and $\texttt{swinb\_cogcoor}$ version of GroundingDINO as the grounding model.
For the RVOS task, we first divide the input video into several clips based on the sampling number and interval of frames.
Then, we conduct pivot frame and box selection within each clip.
All the obtained pivot boxes are used as prompts to SAM 2 while the pivot frame in the middle clip is used as the starting frame of mask propagation.
In each clip, the number of sampled frames is $5$ for all datasets while the sampling interval between frames varies across different datasets.
For the Ref-YouTube-VOS dataset, which already employs a $5$-frame sampling interval, we set an additional sampling interval of $2$ frames, resulting in an actual sampling interval of $10$ frames.
For the Ref-DAVIS17 and MeViS datasets, the sampling interval is set to $5$ frames.
We set GroundingDINO's $\texttt{text\_threshold}$ to $0.2$ and $\texttt{box\_threshold}$ to $0.15$.

For the AVS task, we adopt BEATs~\cite{chen2022beats} as the audio classifier and keep audio categories with the top-$5$ confidence scores in the LBRU module.
Since the video length is relatively shorter in the AVSBench dataset, we do not divide its videos into clips.
For each video in AVSBench, we sample $5$ frames for the S4 and MS3 settings and $10$ frames for the AVSS setting.
We set GroundingDINO's $\texttt{text\_threshold}$ to $0.25$ and $\texttt{box\_threshold}$ to $0.25$.

\subsection{Comparison with State-of-the-Art Methods}
Table~\ref{tab:ytvos_davis} and Table~\ref{tab:avs} present the quantitative comparison results between our method and previous state-of-the-art methods on RVOS and AVS tasks, respectively.
``G-L + SAM 2'' denotes adopting a zero-shot referring image segmentation method Global-Local~\cite{yu2023zero} to obtain the referent mask in the first frame and then using it as the prompt to SAM 2.
As observed in Table~\ref{tab:ytvos_davis}, our proposed training-free \framesimple~achieves the best performance in both training-free and weakly-supervised fine-tuning settings, and it also outperforms most fully-supervised fine-tuning methods on three datasets.
Furthermore, Table~\ref{tab:avs} shows that our method significantly reduces the performance gap between training-free methods and those fine-tuned with full supervision in the AVS task.
These results demonstrate that by designing effective chain-of-thought and unification strategies, we can fully unleash the reasoning and perception capabilities of foundation models, enabling a unified pipeline to empower multiple specific VOS tasks.

\subsection{Ablation Studies}
We conduct ablation studies on the RVOS and AVS tasks to verify the effectiveness of different designs in our pipeline.

\begin{table}[!htbp]
\setlength{\tabcolsep}{4.2pt}
\centering
\footnotesize
\begin{tabular}{r|ccc|ccc}
\rowcolor[gray]{.9}
\hline
& \multicolumn{3}{c|}{Ref-YouTube-VOS} & \multicolumn{3}{c}{Ref-DAVIS17} \\
\rowcolor[gray]{.9} 
Method & \( \mathcal{J} \)\&\( \mathcal{F} \) & \( \mathcal{J} \) & \( \mathcal{F} \)  &  \( \mathcal{J} \)\&\( \mathcal{F} \) & \( \mathcal{J} \) & \( \mathcal{F} \) \\
\hline 
\hline
G-DINO + SAM 2 &64.8 &62.5& 67.0& 66.2 &62.6 &69.7\\
+ PF Select &65.6 &63.6& 67.6& 71.1 &67.0 &74.8\\
+ PF \& PB Select &\textbf{67.9}& \textbf{65.9}& \textbf{69.9}&\textbf{74.2} &\textbf{70.4} &\textbf{78.0}\\
\hline 
\end{tabular}
\caption{Ablation study of the two-step reasoning in our proposed GPT-assisted pivot selection on the Ref-YouTube-VOS and Ref-DAVIS17 datasets.}
\label{tab:ablation_ps_rvos}
\end{table}

\begin{figure*}[!t]
\centering
\includegraphics[width=0.95\linewidth]{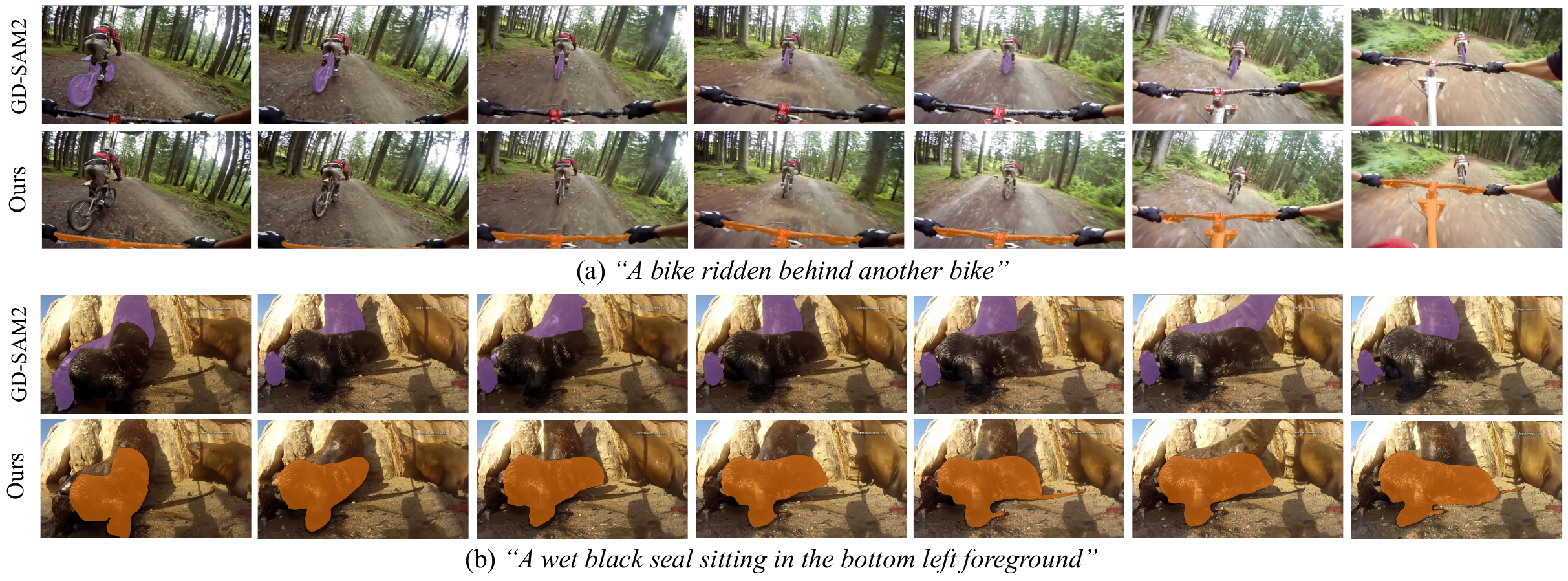}
\caption{Qualitative comparison between our method and the baseline GD-SAM 2 on the Ref-YouTube-VOS dataset.}
\label{fig:vis_rvos}
\end{figure*}

\textbf{Two-step reasoning in GPT-PS.}
As shown in Table~\ref{tab:ablation_ps_rvos}, we validate the effectiveness of the two-step pivot selection process proposed in our GPT-PS on two RVOS datasets.
In the first row, ``G-DINO + SAM 2'' represents our baseline method, which directly uses GroundingDINO on the first frame of the video to obtain the referent's box prediction based on the input language.
This box is then used as a prompt for SAM 2 to obtain the video segmentation results.
In the second row, ``+ PF Select'' denotes integrating the first step of selecting the pivot frame in our GPT-PS into the baseline model.
The last row denotes further integrating the second step of selecting the pivot box upon the first step.
The results show that using only the first step, as well as employing both steps of our proposed GPT-PS, consistently improves the model's performance.
This indicates that leveraging GPT's temporal-spatial reasoning ability to obtain higher-quality prompts can better tap into SAM 2's potential.
Similarly, results in Table~\ref{tab:ablation_lbru_ps_avs} also witness a consistent improvement of both steps in GPT-PS for the AVS task.
This further demonstrates that our proposed temporal-spatial reasoning strategy has excellent generalization ability for various multimodal-referenced VOS tasks under the training-free setting.

\begin{table}[!htbp]
\setlength{\tabcolsep}{5pt}
\centering
\footnotesize
\begin{tabular}{r |c c}
\rowcolor[gray]{.9}
\hline
\rowcolor[gray]{.9} 
Method & \( \mathcal{M_J} \) & \( \mathcal{M_F} \) \\
\hline 
\hline
BEATs Reference & 53.7 & 59.6 \\
LBRU Reference & 56.9 & 64.5  \\
+ PF Selection & 58.4 & 65.9 \\
+ PF \& PB Selection & \textbf{59.2} & \textbf{66.2} \\
\hline 
\end{tabular}
\caption{Ablation study of our proposed LBRU and GPT-PS on the AVSS-V2-Binary dataset.}
\label{tab:ablation_lbru_ps_avs}
\end{table}

\textbf{LBRU reference.}
Furthermore, we also verify the effectiveness of our proposed LBRU on the AVSS-V2-Binary dataset, as shown in the first two rows of Table~\ref{tab:ablation_lbru_ps_avs}.
``BEATs Reference'' denotes that the reference to be fed into GroundingDINO + SAM 2 is generated with the audio categories predicted by BEATs only, which is suboptimal.
Incorporating our proposed LBRU to extract reference achieves notable performance improvement, showing that integrating visual context and using GPT to convert audio signals to language formats can express the audio context more precisely.

\begin{table}[!htbp]
\setlength{\tabcolsep}{5pt}
\centering
\footnotesize
\begin{tabular}{r |c c c}
\rowcolor[gray]{.9}
\hline
& \multicolumn{3}{c}{Ref-YouTube-VOS} \\
\rowcolor[gray]{.9} 
Method & \( \mathcal{J} \)\&\( \mathcal{F} \) & \( \mathcal{J} \) & \( \mathcal{F} \) \\
\hline 
\hline
First Frame & 64.8 & 62.5 & 67.0 \\
Middle Frame & 64.9 & 62.7 & 67.1 \\
Last Frame & 58.5 & 56.5 & 60.5 \\
PF Selection &\textbf{65.6}& \textbf{63.6}& \textbf{67.6}\\
\hline 
\end{tabular}
\caption{Results of different frame selection strategies on the Ref-YouTube-VOS dataset.}
\label{tab:ablation_pfs_rvos}
\end{table}

\textbf{Different frame selection strategies.}
In Table~\ref{tab:ablation_pfs_rvos}, we compare the results of different frame selection strategies on the Ref-YouTube-VOS dataset.
The first three rows represent simple rule-based strategies to select three different fixed frames as the pivot frame.
Our proposed GPT-assisted pivot frame selection achieves the best performance among the four strategies, showing that leveraging GPT's temporal reasoning ability allows us to select a pivot frame that better reflects the reference information, thereby establishing a strong visual context for the entire segmentation.

\begin{table}[!htbp]
\setlength{\tabcolsep}{5pt}
\centering
\footnotesize
\begin{tabular}{r|ccc}
\rowcolor[gray]{.9}
\hline
& \multicolumn{3}{c}{Ref-YouTube-VOS} \\
\rowcolor[gray]{.9}
Method & \( \mathcal{J} \)\&\( \mathcal{F} \) & \( \mathcal{J} \) & \( \mathcal{F} \) \\
\hline
\hline
\textit{w/o} GPT Assisted Box Selection &65.6 &63.6& 67.6 \\
\hline
Select Referent \textit{w/o} Description &63.3 &61.0& 65.5 \\
Describe Then Select &65.6 &63.4& 67.8 \\
\textit{w/} Syntax Analysis of Reference &\textbf{67.9}& \textbf{65.9}& \textbf{69.9}\\
\hline 
\end{tabular}
\caption{Different prompts to GPT for pivot box selection.}
\label{tab:ablation_pbs_rvos}
\end{table}

\textbf{Different box selection prompts to GPT.}
In Table~\ref{tab:ablation_pbs_rvos}, we present the results of using different prompts to GPT in the pivot box selection step to demonstrate the effectiveness of our PF CoT.
The first row means we naively adopt the highest-scoring box predicted by GroundingDINO on the pivot frame as the pivot box without GPT assistance.
The second row indicates that we directly instruct GPT to select the object box that matches the reference information, which instead led to obvious performance degradation, demonstrating that providing GPT with naive instructions makes it even harder to perform correct reasoning.
The third row represents that we first have GPT provide a detailed description of the scene and objects before selecting the pivot box.
Building on the third row, the last row further includes syntax analysis of the language reference, resulting in significant improvement.
These results show that guiding GPT to conduct reasoning with our complete PF CoT prompt benefits the accurate pivot box selection.

\subsection{Qualitative Results}
In Figure~\ref{fig:vis_rvos}, we compare the qualitative results of our method with those of the baseline (GroundingDINO + SAM 2) on the Ref-YouTube-VOS dataset.
These results demonstrate that our method has a stronger ability to distinguish between objects of the same category compared to the baseline.
In the first row, even though only a small part of the bicycle's handlebar is visible, our method still accurately locates the referred bicycle in the back based on cues like ``behind''.

\section{Conclusion}
In this paper, we propose a \framework~(\framesimple) pipeline for training-free audio and language-referenced video object segmentation, namely AVS and RVOS tasks. 
A novel GPT-assisted Pivot Selection (GPT-PS) module is developed to guide GPT-4 for selecting pivot frames and boxes with two-step temporal-spatial reasoning, thereby providing SAM 2 with high-quality initial prompts.
Two task-specific Chain-of-Thought prompts are designed for explicit guidance of GPT within GPT-PS.
Additionally, a Language-Binded Reference Unification (LBRU) module which converts audio signals into language-formatted references is also designed, thus unifying AVS and RVOS tasks within the same pipeline. 
Extensive experiments on both tasks demonstrate that our training-free \framesimple~pipeline achieves results comparable to, or even better than, those of finetuning-based methods.

\section{Acknowledgments}
This research was supported in part by National Science and Technology Major Project (2022ZD0115502), National Natural Science Foundation of China (No. U23B2010, No. 62461160308), Zhejiang Provincial Natural Science Foundation of China (Grant No. LDT23F02022F02), Beijing Natural Science Foundation (No. L231011), Beihang World TOP University Cooperation Program, and Meituan.

\bibliography{aaai25}

\clearpage
\appendix

\section{Details of Datasets}

\textbf{Ref-YouTube-VOS}~\cite{seo2020urvos} is large-scale dataset for referring video object segmentation, which is based on YouTube-VOS dataset~\cite{xu2018youtube}.
It contains 3,975 videos with 7,451 objects and 27,899 referring expressions, where 202 videos are included in the validation set
Two types of annotations are included: full-video expressions, where annotators describe the target object using the entire video, and first-frame expressions, which rely only on the first frame, allowing for the exploration of both static and dynamic properties of the videos. 

\textbf{Ref-DAVIS17}~\cite{khoreva2019video} is an extension of the DAVIS2017 video object segmentation dataset~\cite{pont20172017}, augmented with language descriptions for each annotated object. 
It consists of 60 training videos and 30 validation videos, with multiple objects annotated per video. 
Each video is annotated with pixel-level accuracy, and the dataset includes both first-frame descriptions, where annotators describe the object based on the initial frame, and full-video descriptions, where the entire video is considered. 

\textbf{MeViS}~\cite{ding2023mevis} is a challenging large-scale dataset which contains numerous motion expressions to indicate target objects in complex environments.
It comprises 2,006 videos with a total of 8,171 objects, and 28,570 motion expressions are provided to refer to these objects, where 140 videos are included in the validation split.
It focuses on segmenting objects in videos based on motion cues rather than static attributes, which means target objects cannot be identified by a single frame. 
Instead, motion expressions guide the segmentation, requiring models to capture both fleeting and long-term motions, making this dataset an essential benchmark for developing advanced language-guided video segmentation algorithms.

\textbf{AVS-Bench}~\cite{zhou2023audio} is a large-scale audio-visual segmentation dataset designed for segmenting objects in videos based on the sounds they produce. 
The dataset is divided into three subsets: the Single-source subset (S4), the Multi-sources subset (MS3), and the Semantic-labels subset (AVSS). 
The S4 subset contains 4,932 videos across 23 categories, with only one sounding object exists in each video.
MS3 subset includes 424 videos and involves multiple sound sources. 
he AVSS subset is built upon S4 and MS3 subsets and significantly expands the dataset with 12,356 videos across 70 categories, introducing semantic labels that require the model to generate both segmentation masks and classify the sounding objects by category.
We follow previous work~\cite{wang2024prompting} to conduct segmentation without considering semantic information on the expanded part in AVSS and name the subset as AVSS-V2-Binary accordingly.

\section{Audio Clip Segmentation}
In scenarios where multiple objects may produce sounds simultaneously or intermittently within the same video sample, it is crucial to determine the exact time intervals for each sounding object after obtaining its mask sequence throughout the entire video. 
To this end, we employ the combination of sound event detection (SED) and audio-language feature matching to associate each audio segment with one or more sounding object labels. 
Specifically, we first use a lightweight SED tool\footnote{https://github.com/robertanto/Real-Time-Sound-Event-Detection} to split the audio $\bm{A}$ into $s$ segments $\{\bm{A}^i\}_{i=1}^s$ based on inter-frame feature similarity.
Then, we extract the audio feature for each segment $\bm{f}_{a}^i \in \mathbb{R}^C$ using the audio encoder of LanguageBind~\cite{zhu2023languagebind}, where $C$ denotes the channel number of the audio features.
Let $\{\bm{L}^j\}_{j=1}^d$ be the text labels derived from all $d$ possible combinations of object categories predicted by LBRU, we extract the text feature for each combination as $\bm{f}_t^j \in \mathbb{R}^C$ by feeding the textual labels to the text encoder of LanguageBind.
Finally, since the audio features and the text features are spatially aligned, we compute the similarity between the audio features of each segment and the corresponding text features of the label combinations to associate the audio segment with the corresponding text label combination:
\begin{equation}
l^i = \argmax_{j} {\rm sim}(\bm{f}_a^i, \bm{f}_t^j).
\end{equation}

\section{Detailed Prompts}
\begin{figure*}[!htbp]
\centering
\includegraphics[width=0.7\linewidth]{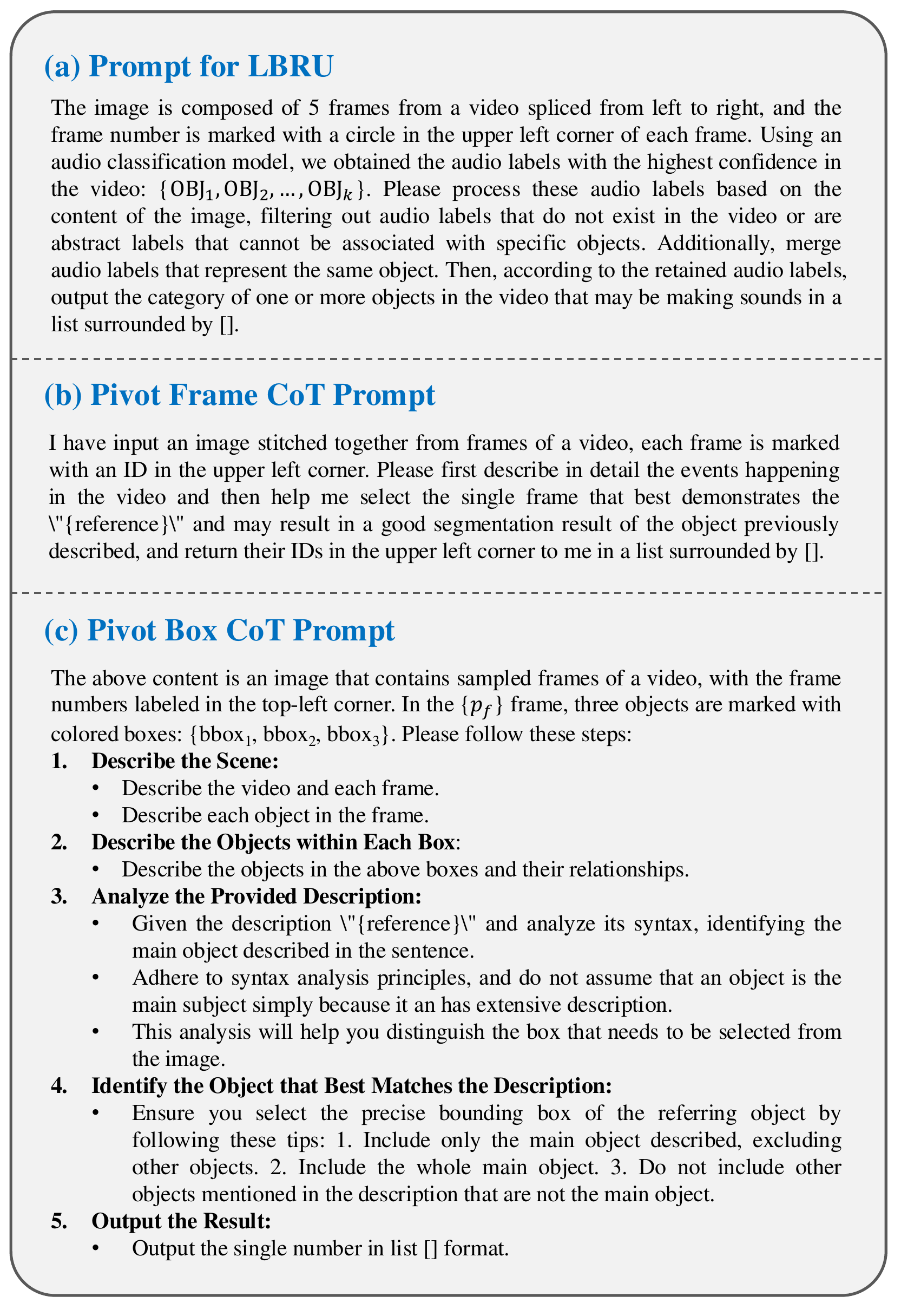}
\caption{Detailed prompts used in our \framesimple~pipeline.}
\label{fig:prompt}
\end{figure*}

We illustrate the detailed version of the prompts used in our \framework~pipeline in Figure~\ref{fig:prompt}.
Since the reference of AVS is much simpler than RVOS, the prompt used in the pivot box selection process for AVS only contains the last two steps in Figure~\ref{fig:prompt}(c).

\section{More Qualitative Results}
\begin{figure*}[!htbp]
\centering
\includegraphics[width=0.95\linewidth]{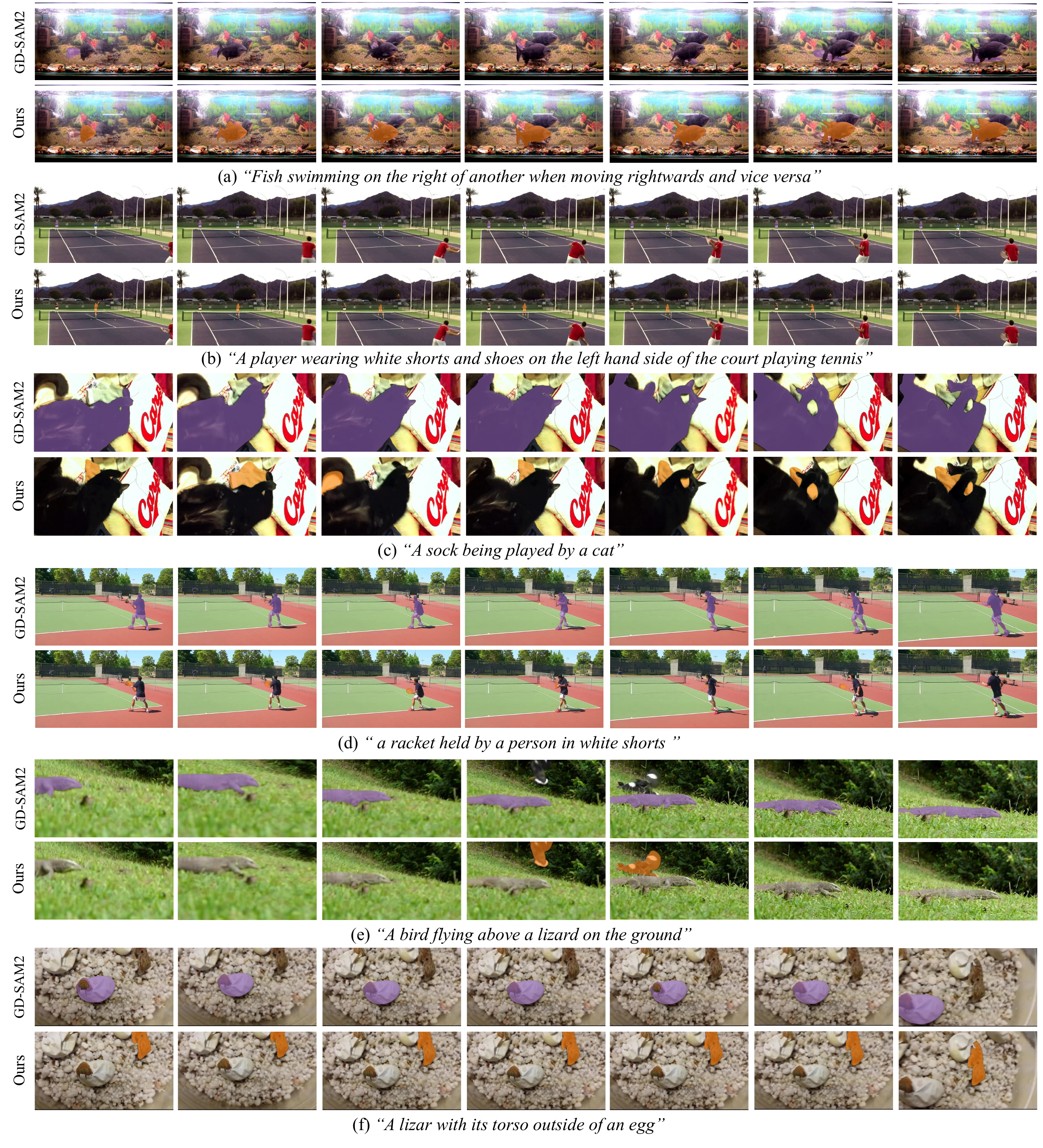}
\caption{Qualitative comparison between our method and the baseline GD-SAM 2 on the Ref-YouTube-VOS dataset.}
\label{fig:vis_rvos2}
\end{figure*}

\begin{figure*}[!htbp]
\centering
\includegraphics[width=0.85\linewidth]{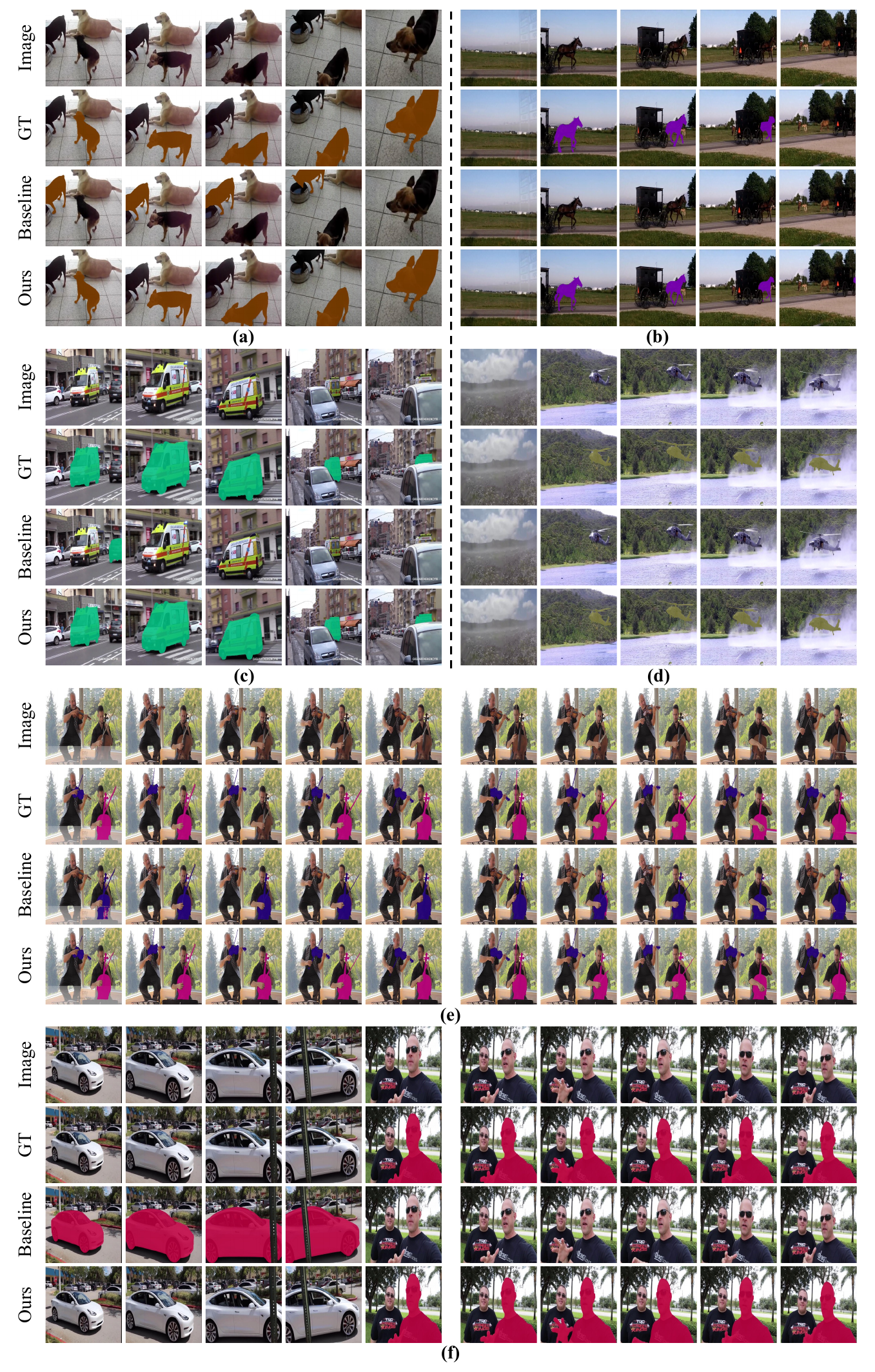}
\caption{Qualitative comparison between our method and the baseline on the AVSS dataset.}
\label{fig:vis_avs}
\end{figure*}

We present more qualitative results on RVOS and AVS tasks in Figure~\ref{fig:vis_rvos2} and Figure~\ref{fig:vis_avs} respectively.
We provide more video demos for the two tasks in the `video\_demos' directory.

\end{document}